# Efficient Characterization of Dynamic Response Variation Using Multi-Fidelity Data Fusion through Composite Neural Network

K. Zhou

Postdoctoral Researcher

J. Tang[†]

Professor

Department of Mechanical Engineering

University of Connecticut

191 Auditorium Road, Unit 3139

Storrs, CT 06269, USA

Phone: (860) 486-5911; Email: jiong.tang@uconn.edu

---

[†] Corresponding author

# Efficient Characterization of Dynamic Response Variation Using Multi-Fidelity Data Fusion through Composite Neural Network

K. Zhou and J. Tang[†]

Department of Mechanical Engineering

University of Connecticut

Storrs, CT 06269, USA

Phone: +1 (860) 486-5911; Email: jiong.tang@uconn.edu

**Abstract**

Uncertainties in a structure is inevitable, which generally lead to variation in dynamic response predictions. For a complex structure, brute force Monte Carlo simulation for response variation analysis is infeasible since one single run may already be computationally costly. Data driven meta-modeling approaches have thus been explored to facilitate efficient emulation and statistical inference. The performance of a meta-model hinges upon both the quality and quantity of training dataset. In actual practice, however, high-fidelity data acquired from high-dimensional finite element simulation or experiment are generally scarce, which poses significant challenge to meta-model establishment. In this research, we take advantage of the multi-level response prediction opportunity in structural dynamic analysis, i.e., acquiring rapidly a large amount of low-fidelity data from reduced-order modeling, and acquiring accurately a small amount of high-fidelity data from full-scale finite element analysis. Specifically, we formulate a composite neural network fusion approach that can fully utilize the multi-level, heterogeneous datasets obtained. It implicitly identifies the correlation of the low- and high-fidelity datasets, which yields improved accuracy when compared with the state-of-the-art. Comprehensive investigations using frequency response variation characterization as case example are carried out to demonstrate the performance.

## 1. Introduction

Engineering structures are usually subject to uncertainties due to material imperfection, manufacturing tolerance, and assemblage error etc. Consequently, their dynamic responses have variations. In order to adequately assess the effect of structural uncertainties, uncertainty propagation analysis of structural dynamic responses becomes an important task. Frequency response function (FRF), as one representative dynamic response, characterizes the fundamental properties of a structure in the frequency domain. As FRF is sensitive to uncertainties especially around resonant frequencies, quantification of FRF variation is commonly involved in robust design and control [1, 2]. One straightforward approach for FRF variation prediction is Monte Carlo simulation through a parametrized, stochastic finite element model [3, 4]. While generally considered accurate when high-fidelity finite element model is employed, a well-known issue of propagating uncertainties from the high-dimensional model to FRF is the high computational cost. For a complex structure, a single run of finite element dynamic analysis may already be costly. Brute force Monte Carlo simulation thus yields prohibitive computational burden [5].

---

[†] Corresponding author

Built upon the rapid advancement in statistical inference, recent exploration of uncertainty quantification of structural dynamic response has focused on various meta-modeling methods that have the prospect of fundamentally reducing the computational cost. The basic idea of meta-modeling such as the Gaussian process is to utilize a small amount of training data, i.e., output response associated with sampled uncertainty parameters, to establish a regression-type relationship of response prediction with respect to the uncertainty parameter set. This can dramatically reduce the number of repetitive simulations or experiments as compared with brute force Monte Carlo methods since the size of training dataset becomes much smaller [6-8]. The meta-model trained can quickly predict frequency responses under given uncertainty parameter samples, which is referred to as emulation. While earlier investigations often resort to single-response meta-model, frequency responses essentially represent the relation of dynamic response of a distributed structure versus excitation frequency and thus feature inherently multiple responses. That is, the frequency responses at different locations have intrinsic correlation and, moreover, the responses at one specific location under different excitation frequencies have intrinsic correlation. In order to account for such correlations and to avoid training multiple meta-models for multiple responses, multi-response Gaussian process (MRGP) technique has been employed which introduces a non-spatial correlation matrix to capture the statistical correlation of different response variables [9-11]. The hyper-parameters dependent on the response correlation are identified through maximizing the multivariate likelihood function. Alternatively, neural network based methods have also been attempted in meta-modeling of structural dynamic response. Actually, neural network can allow directly multi-response emulation through designing an architecture with multiple neurons/nodes at the output layer [12-14]. The correlation of multiple responses can be implicitly established by mutual interaction of different layers. The flexibility and extensibility of neural networks have enabled them to be increasingly used in engineering analysis [8, 15-17].

Intuitively, the performance of a meta-model hinges upon both the quality and quantity of training dataset. That is, larger dataset with high numerical or experimental accuracy is always desired. In actual practice, however, high-fidelity data acquired from high-dimensional finite element simulation or experiment are generally scarce. In order to mitigate this issue, some recent investigations have proposed to incorporate datasets at multiple levels/resolutions to train Gaussian process meta-model. It was suggested that the combination of data with different fidelities for Gaussian process emulation could maintain both prediction accuracy and efficiency [18-20]. In the realm of structural dynamic analysis, a natural way of carrying out fast, low-fidelity simulation is through reduced-order modeling [5, 21]. A large amount of first principle-based simulation data can be produced easily with reduced-order model and then employed in the multi-level Gaussian process meta-modeling. At the same time, a small amount of high-fidelity, full-scale finite element simulation data will also be employed in the training. With the large amount of low-fidelity data, the Gaussian process emulator can avoid those errors associated with the inference procedure. With the introduction of a few high-fidelity data, we can correct the error of the low-fidelity data inherited from the order-reduction procedure. The advantage of such a heterogeneous data-driven meta-modeling that combines low- and high-fidelity datasets has been demonstrated in structural vibration analysis case [22]. Since structural dynamic responses are generally characterized in a distributed manner, the above-mentioned multi-response Gaussian process has recently been extended to multi-level and multi-response Gaussian process (MLMRGP) that is capable of emulating distributed outputs (i.e., predicting multiple output variables simultaneously) [23]. One challenging issue in these Gaussian process trainings is the overall training cost and in particular numerical stabilities in large-scale matrix operations. Another intriguing issue in multi-level Gaussian process meta-modeling is the treatment of the correlation between datasets at different levels/resolutions. In [22], it was assumed that there was linear correlation which was characterized by a linear autoregressive scheme. Perdikaris et al [24] however argued that in such fusion between multi-level datasets, nonlinear correlation should be learned which requires complex, additional computational efforts.

Indeed, structural dynamic systems offer interesting potentials that could be leveraged upon to yield efficient and accurate uncertainty quantification. One may formulate multi-level analyses to produce multi-level datasets of frequency responses. The responses of a structure are distributed in nature, leading to multiple outputs with intrinsic correlations. In view of the prospect of meta-modeling as well as the pros and cons of methods developed so far, in this research we explore a new framework of frequency response variation characterization built upon the neural network concept. Neural networks can accommodate both classification and regression based on supervised learning, and possess flexible architectures. Uncertainty quantification of dynamic responses falls into the category of regression. Neural networks generally are capable of providing multiple outputs. Recent progresses indicate that it is possible to synthesize a composite neural network that can fuse together heterogeneous data to facilitate learning, known as multi-fidelity physics-informed neural network (MFPINN) [25]. Our hypothesis here is that, exploiting the architecture of such heterogeneous data-driven neural network and its inherent learning capability, we can establish a new path toward multi-level meta-modeling. Specifically, the proposed new approach, hereafter referred to as multi-fidelity data fusion composite neural network (MFDF-CNN), will use a large amount of low-fidelity data produced by reduced-order analysis, and a small amount of high-fidelity data generated by full-scale finite element model. The key advancement is that this MFDF-CNN will feature the built-in function of incorporating the correlation between low- and high-fidelity datasets used in training, thereby addressing the current issue in multi-level meta-modeling. In other words, the network can synergistically fuse multi-fidelity datasets in an integral manner, yielding improved accuracy in uncertainty quantification of frequency response.

The rest of this paper is organized as follows. Section 2 outlines the generation of high- and low-fidelity datasets for frequency response analysis. In Section 3, we start from presenting a multi-level multi-response Gaussian process (MLMRGP) for frequency response emulation, and then formulate the proposed multi-fidelity data fusion composite neural network (MFDF-CNN) for meta-modeling. Through direct comparison, the advantage of MFDF-CNN is elaborated. Section 4 provides comprehensive case studies to demonstrate the new methodology and the improved performance. Section 5 summarizes the concluding remarks.

## 2. Problem Setup and Multi-Fidelity Datasets Generalization

### 2.1. *High-fidelity data generation through full-scale finite element frequency response analysis*

We assume the full-scale finite element model of a structure is available, i.e.,

$$\mathbf{M}(\boldsymbol{\theta})\ddot{\mathbf{z}} + \mathbf{C}(\boldsymbol{\theta})\dot{\mathbf{z}} + \mathbf{K}(\boldsymbol{\theta})\mathbf{z} = \mathbf{f} \qquad (1)$$

where $\mathbf{M}(\boldsymbol{\theta})$, $\mathbf{C}(\boldsymbol{\theta})$, and $\mathbf{K}(\boldsymbol{\theta})$ are, respectively, the mass, damping, and stiffness matrices with dimension $N \times N$, and $\boldsymbol{\theta}$ is an *m*-dimensional vector representing the set of *m* uncertain parameters in the model. *N* is the number of degrees of freedoms (DOFs), **z** is the displacement vector, and **f** is the external excitation vector. Without loss of generality, we assume proportional damping. The uncertainties in the structural model yield the variation of the response. In this research, we are specifically interested in frequency response of the structure. Let us consider a harmonic excitation $\mathbf{f}(t) = \mathbf{F}e^{j\omega t}$ where **F** is a constant vector of force magnitude and $\omega$ is the sweeping frequency. We then have the vector-form frequency response function of the structure as

$$\mathbf{Z}(\boldsymbol{\theta}) = [-\omega^2 \mathbf{M}(\boldsymbol{\theta}) + j\omega \mathbf{C}(\boldsymbol{\theta}) + \mathbf{K}(\boldsymbol{\theta})]^{-1}\mathbf{F} \qquad (2)$$

where $\mathbf{Z}(\boldsymbol{\theta})$ is the vector-form response amplitude of the entire structure, and subjected to variations owing to structural uncertainties.

In actual practice, usually only the responses at a selected number of DOFs are of interest for design and control applications. Meanwhile, experimental results can only be acquired at a small number of locations due to the usual constraint in the number of sensors. Therefore, hereafter we analyze **U**, a subset of **Z**. To begin with, **U** is an *n*-dimensional vector ($n < N$) and is dependent upon $\omega$. We further assume that frequency responses at pre-specified *p* discrete frequency points, $\boldsymbol{\omega} = [\omega_1, \omega_2, \cdots, \omega_p]$, are of interest. For simplicity in notation, the responses of the structure at these *n* DOFs and *p* frequency points are collectively expressed as

$$\mathbf{U}(\boldsymbol{\theta}) = [U_{1,1}, \cdots U_{n,1}, U_{1,2}, \cdots, U_{n,2}, \cdots\cdots, U_{1,p}, \cdots, U_{n,p}]^T \tag{3}$$

Obviously, **U** is dependent upon uncertainty parameter set $\boldsymbol{\theta}$. In order to avoid brute force Monte Carlo simulation that leads to prohibitive computational cost, we resort to meta-modeling for the uncertainty quantification of frequency response **U**. High-fidelity data can be produced from Equation (2) (i.e., full-scale finite element model) directly under sampled uncertainty parameter set $\boldsymbol{\theta}$. In actual practice, the amount of high-fidelity data is usually limited, due to the computational cost involved in full-scale finite element analysis.

## 2.2. Low-fidelity data generation through reduced-order frequency response analysis

Since dynamic analysis of high-dimensional finite element model is computationally costly, model order reduction has been one important research subject is structural dynamic analysis. A variety of approaches have been proposed in recent decades. The goal of this research is to establish a new meta-modeling approach that can integrally utilize a small amount of high-fidelity data together with a large amount of low-fidelity data. Specifically, we hope the usage of large amount of low-fidelity data, to be produced by a reduced-order model, can mitigate the error associated with the inference procedure. Meanwhile, we hope that the introduction of a small amount of high-fidelity data based on the preceding subsection can correct the error of the low-fidelity data inherited from the order-reduction procedure. Without loss of generality and in order to make it easy for interested readers to re-produce the case investigation, here we adopt the Guyan reduction for reduced-order modeling [26].

In Guyan reduction, the DOFs in the finite element model are divided into the master DOFs and the slave DOFs. The effects of the slave DOFs are transformed onto the master DOFs through static condensation, thereby eliminating the slave DOFs in the original model. We thus re-write the equation of motion of as

$$\begin{bmatrix} \mathbf{M}_{mm} & \mathbf{M}_{ms} \\ \mathbf{M}_{sm} & \mathbf{M}_{ss} \end{bmatrix} \begin{bmatrix} \ddot{\mathbf{z}}_m \\ \ddot{\mathbf{z}}_s \end{bmatrix} + \begin{bmatrix} \mathbf{C}_{mm} & \mathbf{C}_{ms} \\ \mathbf{C}_{sm} & \mathbf{C}_{ss} \end{bmatrix} \begin{bmatrix} \dot{\mathbf{z}}_m \\ \dot{\mathbf{z}}_s \end{bmatrix} + \begin{bmatrix} \mathbf{K}_{mm} & \mathbf{K}_{ms} \\ \mathbf{K}_{sm} & \mathbf{K}_{ss} \end{bmatrix} \begin{bmatrix} \mathbf{z}_m \\ \mathbf{z}_s \end{bmatrix} = \begin{bmatrix} \mathbf{f}_m \\ \mathbf{f}_s \end{bmatrix} \tag{4}$$

where subscripts $m$ and $s$ denote the master and slave DOFs, respectively. Neglecting the inertia and damping terms and assuming free vibration without external excitation, we can obtain the following approximate relation between the slave and master DOFs,

$$\mathbf{z}_s = -\mathbf{K}_{ss}^{-1}\mathbf{K}_{sm}\mathbf{z}_m \tag{5}$$

which yields

$$\begin{bmatrix}\mathbf{z}_m \\ \mathbf{z}_s\end{bmatrix} = \begin{bmatrix}\mathbf{I} \\ -\mathbf{K}_{ss}^{-1}\mathbf{K}_{sm}\end{bmatrix}\mathbf{z}_m = \mathbf{T}_G\mathbf{z}_m \tag{6}$$

where $\mathbf{T}_G$ is the condensation transformation matrix for Guyan reduction. Utilizing the coordinate transformation shown above, we have

$$\mathbf{M}_r\ddot{\mathbf{z}}_m + \mathbf{C}_r\dot{\mathbf{z}}_m + \mathbf{K}_r\mathbf{z}_m = \mathbf{f}_r \tag{7}$$

where $\mathbf{M}_r = \mathbf{T}_G^T\mathbf{M}\mathbf{T}_G$, $\mathbf{C}_r = \mathbf{T}_G^T\mathbf{C}\mathbf{T}_G$, $\mathbf{K}_r = \mathbf{T}_G^T\mathbf{K}\mathbf{T}_G$ and $\mathbf{f}_r = \mathbf{T}_G^T\mathbf{f}$ are the order-reduced mass, damping, stiffness matrices and the corresponding order-reduced external excitation, respectively. Assuming harmonic excitation $\mathbf{f}_r = \mathbf{F}_r e^{j\omega t}$ where $\mathbf{F}_r$ is a constant vector of force magnitude and $\omega$ is the sweeping frequency, we obtain the frequency response function of the reduced-order system,

$$\mathbf{Z}_m = [-\omega^2\mathbf{M}_r + j\omega\mathbf{C}_r + \mathbf{K}_r]^{-1}\mathbf{F}_r \tag{8}$$

where $\mathbf{Z}_m$ is the vector-form response amplitude of the reduced-order system. Recall Equation (6). The vector-form response amplitude of the original model can be obtained as

$$\mathbf{Z} = \begin{bmatrix}\mathbf{Z}_m \\ \mathbf{Z}_s\end{bmatrix} = \begin{bmatrix}\mathbf{I} \\ -\mathbf{K}_{ss}^{-1}\mathbf{K}_{sm}\end{bmatrix}\mathbf{Z}_m = \mathbf{T}_G\mathbf{Z}_m \tag{9}$$

The system coefficient matrices and the response vectors are all subjected to variations and uncertainties, and thus are $\boldsymbol{\theta}$ dependent. For notation simplicity, we have omitted $\boldsymbol{\theta}$ in the above equations.

Once again, we assume only the responses at a selected number of $n$ DOFs are of interest, and furthermore the frequency responses at pre-specified $p$ discrete frequency points, $\boldsymbol{\omega} = [\omega_1, \omega_2, \cdots, \omega_p]$, are acquired. The responses of the structure at these $n$ DOFs and $p$ frequency points, obtained through reduced-order finite element-based simulations, are collectively expressed as

$$\mathbf{U}_r(\boldsymbol{\theta}) = [U_{r1,1}, \cdots U_{rn,1}, U_{r1,2}, \cdots, U_{rn,2}, \cdots\cdots, U_{r1,p}, \cdots, U_{rn,p}]^T \tag{10}$$

Indeed, vector $\mathbf{U}_r$ corresponds to $\mathbf{U}$ shown in Equation (3), whereas the subscript $r$ indicates reduced-order analysis result. As the dimension of the finite element-based model is reduced from $N$ to $N_m$ (the number of master DOFs) after Guyan reduction, the computational cost involved in Equation (10) is significantly reduced. We may conduct frequency response simulations with a large sample size of uncertainty parameters to acquire a large amount of low-fidelity training data. Nevertheless, the order-reduction introduces truncation errors because the inertia and damping effects of the slave DOFs are neglected in the transformation (Equations (5) and (6)).

## 3. Multi-fidelity Data Fusion Composite Neural Network for Response Variation Characterization

In this section, we outline the new computational framework of multi-fidelity data fusion composite neural network (MFDF-CNN) for meta-modeling to facilitate efficient and accurate uncertainty quantification of frequency responses of structures. We intend to employ a small amount of high-fidelity data generated by full-scale finite element analysis (Section 2.1) and a large amount of low-fidelity data generated by reduced-order analysis (Section 2.2) to train the meta-model. As will be shown, the new approach exploits the architecture of heterogeneous data-driven neural network and its inherent learning capability, and can overcome certain shortcoming of multi-level multi-response Gaussian process approach. For comparison purpose and to highlight the improvement, we start from outlining a multi-level multi-response Gaussian process (MLMRGP) for frequency response emulation, and then formulate the proposed composite neural network. Both methods will be tested in the subsequent case investigations.

### 3.1. Response variation emulation using MLMRGP as baseline

Gaussian processes based meta-modeling has seen wide applications especially in uncertainty quantification. The underlying idea is to extend the multivariate Gaussian distribution from a finite dimensional space to an infinite dimensional space. It essentially yields a probabilistic framework for nonparametric regression. Commonly used Gaussian processes include multi-response regression with single-type training dataset or single-response regression with multi-fidelity datasets [9-11, 18]. In a recent effort, a multi-level multi-response Gaussian process (MLMRGP) meta-model was attempted to integrate multi-fidelity datasets to emulate multi-responses in the uncertainty quantification of dynamic responses [23]. The mathematical foundation is briefly outlined here.

In Gaussian process formulation, an unknown system is denoted as $g(\mathbf{x})$, in which $\mathbf{x}$ is an input vector. In the context of uncertainty quantification discussed in this research, the input vector is the sample of the set of uncertainty parameters $\boldsymbol{\theta}$ mentioned in Section 2. We aim at finding the best $g(\mathbf{x})$ such that $g(\mathbf{x}) \approx \mathbf{y}$, where $\mathbf{y}$ is an output vector or the frequency response vector $\mathbf{U}$ (i.e., multi-responses) mentioned in Section 2, through utilizing training dataset(s). In MLMRGP, we use two-level datasets, i.e., low- and high-fidelity datasets in training. They are denoted as $\vartheta^{(u)} = \{(\mathbf{y}_i^{(u)}, \mathbf{x}_i^{(u)}), i=1,2,\dots n_s^{(u)}; u=1,2\}$, where superscript $u$ indicates the $u$-th data level and $n_s^{(u)}$ is the number of the data points. The dimension of $\mathbf{x}_i^{(u)}$ (i.e., $\boldsymbol{\theta}$) is $m$, and the dimension of $\mathbf{y}_i^{(u)}$ is $n \times p$. Here $\vartheta^{(2)}$ is the low-fidelity dataset ($\mathbf{U}_r$ shown in Equation (10)) produced by the reduced-order frequency response analysis, and $\vartheta^{(u)}$ is the high-fidelity dataset ($\mathbf{U}$ shown in Equation (3)) produced by the full-scale finite element frequency response analysis.

We assume a quasi-linear relation between the low- and high-fidelity outputs, expressed with an autoregressive scheme [18],

$$\mathbf{y}^{(2)} = \rho^{(1)}\mathbf{y}^{(1)} + \boldsymbol{\delta}^{(2)} \tag{11}$$

where $\rho^{(1)}$ is a regression variable. $\mathbf{y}^{(1)}$ and $\boldsymbol{\delta}^{(2)}$ are two independent stationary multivariate Gaussian processes. As the summation of independent Gaussians remains in the closed form, we can derive the Gaussian process representation of observed low- and high-fidelity data points as

$$\begin{bmatrix} \mathbf{y}^{(1)} \\ \mathbf{y}^{(2)} \end{bmatrix} \sim \text{GP}(\mathbf{h}(\mathbf{x})\boldsymbol{\beta}, \mathbf{Q}\boldsymbol{\Sigma}(\mathbf{x},\mathbf{x}')) \tag{12}$$

The first item at the right-side of Equation (12), $\mathbf{h}(\mathbf{x})\boldsymbol{\beta}$, represents the linear mean functions of all outputs. $\boldsymbol{\Sigma}(\mathbf{x},\mathbf{x}')$ is the spatial covariance matrix. Each entry of $\boldsymbol{\Sigma}(\mathbf{x},\mathbf{x}')$ is a value evaluated with the covariance function/kernel, describing the behavior of the process regarding the separation of any two input points. In this research, a commonly adopted covariance function/kernel, i.e., the squared exponential function $\Sigma_{ij}^{(u)} = \exp\left\{-\sum_{k=1}^{r} b_k^{(u)}\left(x_{i,k} - x_{j,k}\right)^2\right\}$, is used. $\mathbf{Q}$ is non-spatial correlation matrix that is intended to characterize the internal correlation among multiple output variables.

The training process follows the Bayesian framework that aims at maximizing the likelihood formulated in terms of the training datasets. The likelihood is expressed as

$$p(\mathbf{y}^{(2)^*} \mid \mathbf{x}^{(1)}, \mathbf{y}^{(1)}, \mathbf{x}^{(2)}, \mathbf{y}^{(2)}, \mathbf{x}^{(2)^*}, \boldsymbol{\varphi}) = \frac{p(\mathbf{y}^{(2)^*} \mid \mathbf{x}^{(2)^*}, \boldsymbol{\varphi}) p(\mathbf{y}^{(1)}, \mathbf{y}^{(2)} \mid \mathbf{x}^{(1)}, \mathbf{x}^{(2)}, \mathbf{x}^{(2)^*}, \mathbf{y}^{(2)^*}, \boldsymbol{\varphi})}{p(\mathbf{y}^{(1)}, \mathbf{y}^{(2)} \mid \mathbf{x}^{(1)}, \mathbf{x}^{(2)}, \boldsymbol{\varphi})} \tag{13}$$

$\boldsymbol{\varphi}$ represents the hyper-parameters from the mean and covariance functions to be optimized. The inputs of the high-fidelity dataset should be a subset of that of the low-fidelity dataset in order to facilitate the emulation. The low- and high-fidelity datasets can be sequentially plugged into training process. Once training is completed, the high-level prior will be updated to the posterior with optimized hyper-parameters,

$$\left[\mathbf{y}^{(2)^*}\right] \sim \text{GP}(\mathbf{h}(\mathbf{x}^{(2)^*})\hat{\boldsymbol{\beta}}, \hat{\mathbf{Q}}\hat{\boldsymbol{\Sigma}}(\mathbf{x}^{(2)^*},\mathbf{x}^{(2)^*})) \tag{14}$$

where $\hat{\boldsymbol{\beta}}$, $\hat{\mathbf{Q}}$ and $\hat{\boldsymbol{\Sigma}}$ are coefficients of the updated mean function, and the updated non-spatial and spatial covariance functions in terms of the optimized hyper-parameters $\hat{\boldsymbol{\varphi}}$.

### 3.2. Response variation emulation using MFDF-CNN

Owing to the advancements in computational power and data science, machine learning through neural network has seen rapid progress in recent years. In this subsection we outline the architecture of a composite neural network specifically tailored toward multi-fidelity data training for frequency response variation emulation.

The basic unit of a neural network is neuron, or node [27]. Its function is to compute the output based on the input received from other nodes, as illustrated in Figure 1. As can be seen, each input $x_i$ has its corresponding weight $\omega_i$, and each node has one bias $b_j$. A nonlinear activation function $f$ is applied onto the linear weighted sum to yield output of node $y_j$. The purpose of the activation function is to introduce inherent non-linearity of the input-output relation into the process. Frequently employed activation functions include sigmoid $1/(1+e^{-x})$, hyperbolic tangent $(e^x - e^{-x})/(e^x + e^{-x})$, and ReLU expressed as $f(x) = \max(0, x)$. A neural network essentially is built upon different layers such as input layer, hidden layer, and output layer, by linking nodes. Hidden layers undertake the major computation to extract underlying data features. According to the respective configuration, hidden layers can be further divided into fully connected layers, convolutional layers, and max pooling layers. While fully connected layers are widely utilized, convolutional layers and max pooling layers are oftentimes integrated into deep learning convolutional neural networks to deal with large amount of training data [28-30]. In terms of node connection pattern, there exist feedforward neural networks and recurrent neural networks [31]. The key step in neural network training is to identify the weights and biases through learning, e.g., minimizing a cost/loss function using training dataset,

$$Loss = \frac{1}{s}\sum_j (y_j - \sum_i (\omega_i x_i + b_j))^2 \qquad (15)$$

where s is the number of training data. Back propagation optimization is typically used [31]. In general, the architecture of a neural network can be quite flexible.

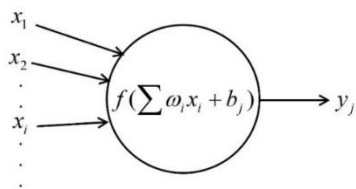

Figure 1. A single node in a neural network

In this research we aim at advancing the meta-modeling of frequency response variation analysis utilizing multi-fidelity datasets. An assumption made in MLMRGP meta-modeling outlined in the preceding subsection is that the autoregressive correlation between the low- and high-fidelity outputs is linear (Equation (11)), which has been adopted in similar investigations [18, 22]. In reality, however, there is no guarantee that such correlation is linear. In fact, generally, the relation between the low- and high-fidelity outputs should be re-written as

$$\mathbf{y}^{(2)} = v(\mathbf{y}^{(1)}, \mathbf{x}) \tag{16}$$

where $v(\cdot)$ is an unknown function that reflects the implicit relation between the low- and high-fidelity outputs. Obviously, one may not be able to solve this problem using Gaussian processes, since only explicitly linear correlation can maintain $\mathbf{y}^{(2)}$ as a Gaussian distribution. To solve this fundamental issue, hereafter we resort to the neural network approach owing to its flexibility in architecture design and customization. In particular, we develop a multi-fidelity data fusion composite neural network (MFDF-CNN) that is capable of taking the implicit relation between the low- and high-fidelity outputs into account. The architecture of MFDF-CNN is shown in Figure 2. The rationale of this proposed approach is outlined as follows. Algorithmic details will be further discussed in the subsequent section through implementing to frequency response variation prediction.

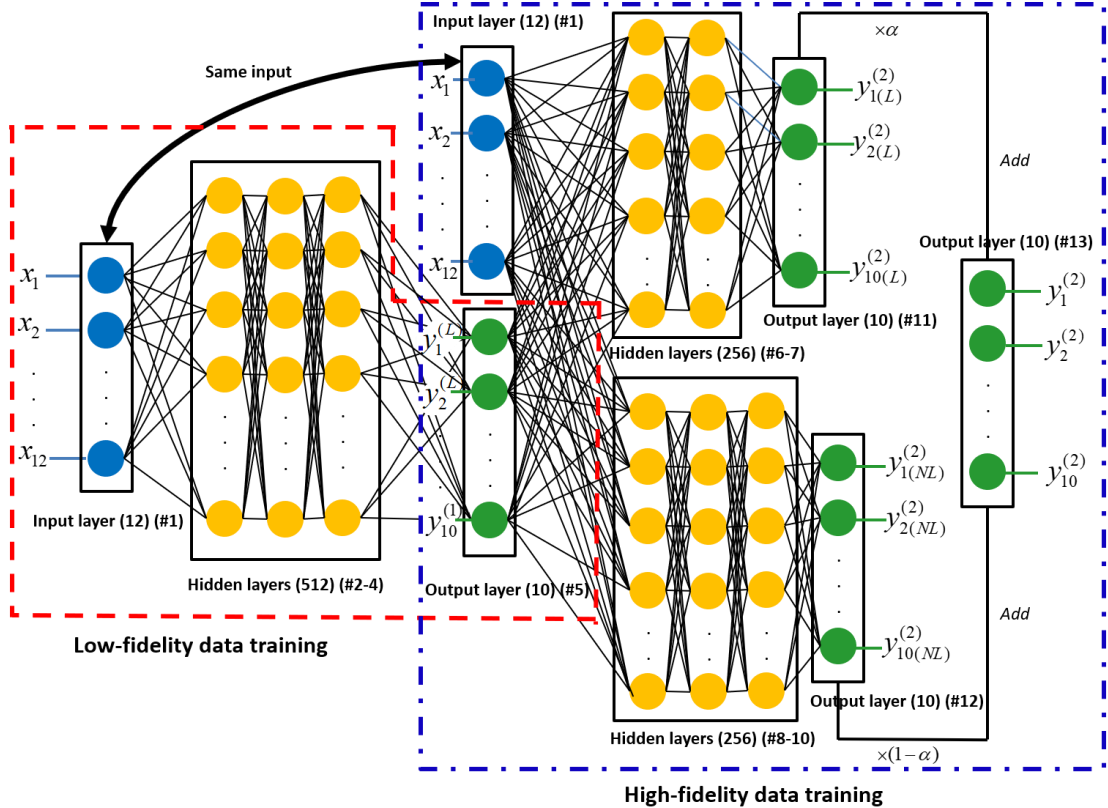

Figure 2. Architecture of multi-fidelity data fusion composite neural network (MFDF-CNN).

We start from the model training process. As we aim at emulating the frequency response of the structural system with uncertainties, the input of this model is the sampled set of uncertainty parameters. Two outputs are produced, including the low-fidelity response in the middle of Figure 2 and the high-fidelity response at the rightmost side of Figure 2. First, the model instance is randomly initialized by assigning arbitrary weights and biases to all layers. The performance of the initial model is then checked by forward propagation analysis, including two sequential procedures. Let us consider one data point, i.e., the $i$-th data point $\mathbf{d}_i = (\mathbf{x}_i, \mathbf{y}_i^{(1)}, \mathbf{y}_i^{(2)})$, that is passed into the model. Here, $\mathbf{x}_i$ is the $i$-th set of uncertainty parameters, and $\mathbf{y}_i^{(1)}$ and $\mathbf{y}_i^{(2)}$ are the corresponding low- and high-fidelity frequency response vectors. In the first procedure (boxed by dashed lines in Figure 2), the goal is to facilitate the prediction of low-fidelity output under current input, i.e., $\mathbf{x}_i$, which thus yields the low-fidelity propagation error with respect to the actual low-fidelity output, i.e., $\mathbf{y}_i^{(1)}$. In the second procedure (boxed by dash-dotted lines in Figure 2), we let the same input, together with the predicted low-fidelity output, further propagate through the rest of the model (i.e., a number of hidden layers). This essentially realizes the characterization of implicit relation shown in Equation (16). Indeed, we decompose such implicit relation into the linear and nonlinear parts with associated weights in the second procedure, expressed as [25]

$$\tilde{\mathbf{y}}_i^{(2)} = \alpha v_L(\mathbf{y}_i^{(1)}, \mathbf{x}_i) + (1-\alpha) v_{NL}(\mathbf{y}_i^{(1)}, \mathbf{x}_i), \qquad \alpha \in [0,1] \tag{17}$$

where $\alpha$ is the weight of the linear part, i.e., linearity weight. Subscripts $L$ and $NL$ indicate the linear and nonlinear parts, respectively. $\tilde{\mathbf{y}}_i^{(2)}$ represents the predicted high-fidelity output through forward propagation analysis. Specifically, in the second procedure the propagation proceeds along two parallel passages. The hidden layer ensemble with linear activation functions at the top passage represents an implicitly linear function, i.e., $v_L(\mathbf{y}_i^{(1)}, \mathbf{x}_i)$, which is to learn the linear behavior within data. Multiplying $v_L(\mathbf{y}_i^{(1)}, \mathbf{x}_i)$ by a fraction, i.e., the linearity weight $\alpha$, hence indicates the portion of the resulting high-fidelity output, i.e., $\tilde{\mathbf{y}}_i^{(2)}$, that is linearly associated with the input and the low-fidelity output. Likewise, $v_{NL}(\mathbf{y}_i^{(1)}, \mathbf{x}_i)$ located at the bottom passage is an implicitly nonlinear function by specifying nonlinear activation functions in the hidden layers. Accordingly, the nonlinear portion of the resulting high-fidelity output, i.e., $\tilde{\mathbf{y}}_i^{(2)}$ can be characterized as $(1-\alpha) v_{NL}(\mathbf{y}_i^{(1)}, \mathbf{x}_i)$. It is worth pointing out that $v_L(\mathbf{y}_i^{(1)}, \mathbf{x}_i)$ and $v_{NL}(\mathbf{y}_i^{(1)}, \mathbf{x}_i)$ are mathematically described with related weights and biases in the hidden layers. To accurately construct $v_L(\mathbf{y}_i^{(1)}, \mathbf{x}_i)$ and $v_{NL}(\mathbf{y}_i^{(1)}, \mathbf{x}_i)$, those layer weights and biases need to be optimized through training process. The high-fidelity output finally can be predicted by merging the information in two passages following Equation (17). By comparing the predicted high-fidelity output with the actual high-fidelity output, i.e., $\mathbf{y}_i^{(2)}$, the high-fidelity propagation error can be calculated.

As mentioned, there are two types of propagation errors generated. To take advantage of both low- and high-fidelity datasets, both errors should be taken into account during the training. However, neural network training generally is performed upon one single loss in order to concurrently optimize all inter-related weights and biases. One direct solution is to properly aggregate different errors together into one. Therefore, in this study we let the initial loss of the *i*-th data point be expressed as a combination of the low-fidelity propagation error and the high-fidelity propagation error, i.e.,

$$\eta_i = \gamma \left\| \mathbf{y}_i^{(1)} - \tau(\mathbf{x}_i) \right\| + (1-\gamma) \left\| \mathbf{y}_i^{(2)} - v(\mathbf{x}_i, \tau(\mathbf{x}_i)) \right\| \tag{18}$$

where $\tau(\cdot)$ is an implicit function that characterizes the relation between the low-fidelity input and output. $\gamma$ is the weight of loss contributed by the low-fidelity output, and $(1-\gamma)$ is the weight of loss contributed by the high-fidelity output. $v(\cdot)$ is the high-level emulator shown in Equation (16), and $\|\cdot\|$ is the Euclidean distance employed to measure the propagation errors of two output vectors. The weights of linearity and loss are the hyper-parameters of MFDF-CNN, and are tuned based on output data characteristics empirically. In certain cases, one may optimize these hyper-parameters through grid search [32].

In training MFDF-CNN, we aim at minimizing the total loss or loss function of all training data. In this research, we adopt the mean squared error (MSE) given as

$$\eta = \frac{1}{n_s} \sum_{i=1}^{n_s} \eta_i^2 \tag{19}$$

where $n_s$ is the number of training data points. Once all training data are introduced, the back propagation optimization will be utilized to update the weights and biases of the model iteratively until the loss function reaches the minimum. It is still worth noting that the proposed architecture (Figure 2) is intended to utilize the low- and high-fidelity data in a carefully designed, inter-related parallel manner. If additional levels of data are available, we can easily generalize this hierarchical architecture to handle more than two data fidelities. This can be achieved by further incorporating sequential and interactive layer ensembles similar to the boxed ones in Figure 2.

## 4. Algorithmic Details and Implementation

In this section, we first produce high-fidelity and low-fidelity data of frequency responses from a benchmark plate structure. We then implement multi-fidelity data fusion to establish meta-models using MLMRGP (the baseline) and MFDF-CNN (the proposed method), respectively. Our focus is on the algorithmic details of MFDF-CNN as well as its advantages over MLMRGP.

### *4.1. Benchmark structure setup and data preparation*

We analyze a benchmark plate structure (Figure 3) for case demonstration. The mass density, Young's modulus and Poisson's ratio of this plate are $7.85 \times 10^3$ kg/m$^3$, 206 GPa and 0.3, respectively. Proportional damping, i.e., $\mathbf{C} = a_M \mathbf{M} + a_K \mathbf{K}$ is used in Equations (1) and (4), where $a_M$ and $a_K$ are 0.01 and 0.0001, respectively. We use 8-node solid element in discretization. The finite element model has 3,510 DOFs.

We choose this structural configuration so interested readers can readily re-construct the mesh for validation and comparison. As can be observed, this structure consists of three smaller plates joined together, which resembles topologies of complex engineering structures consisting of multiple substructures.

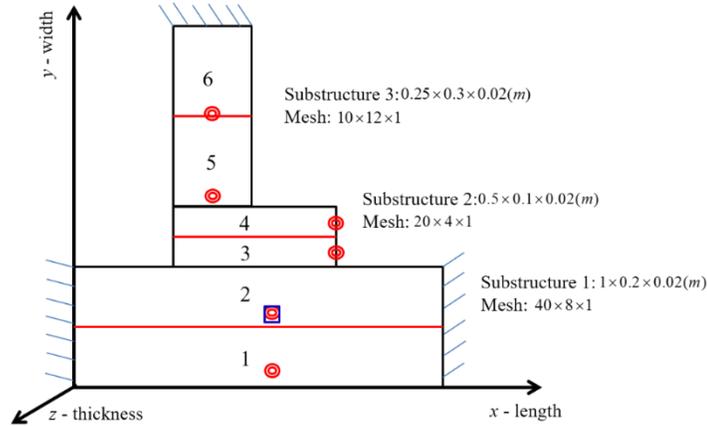

Figure 3. Benchmark plate structure. ⊚ and ☐ indicate location where harmonic unit force is applied and where response is of interest.

**Table 1. Comparison of natural frequencies between full-scale finite element analysis and Guyan reduced-order analysis (Hz)**

| Mode Order | Full-Scale Analysis | Guyan Reduction |
|---|---|---|
| 1 | 144.3078 | 144.5716 |
| 2 | 334.7630 | 345.8223 |
| 3 | 367.8749 | 373.9271 |
| 4 | 571.4485 | 596.7220 |
| 5 | 709.2347 | 828.6293 |

In this research, the result of full-scale finite element analysis (Section 2.1) is referred to as the high-fidelity data. A reduced-order model using Guyan reduction (Section 2.2) is developed and employed to generate the low-fidelity data. The reduced-order model has 230 DOFs, and thus is computationally efficient. On the other hand, since inertia effects of slave DOFs are omitted, the reduced-order results are subject to error. A comparison of the natural frequencies computed from the full-scale model and the Guyan reduced-order model is given in Table 1. Generally, the reduced-order model yields larger error for higher order natural frequencies. We focus on frequency responses in this study for case demonstration. As indicated in Figure 3, frequency sweeping harmonic forces with unit amplitude are applied at 6 locations/DOFs. We are interested in the response amplitudes at one of these locations/DOFs. Specifically, we pick a total of 10 frequency points that are uniformly discretized from 120 Hz to 170 Hz, i.e., 120 Hz, 125.56 Hz, 131.11 Hz, 136.67 Hz, 142.22 Hz, 147.78 Hz, 153.33 Hz, 158.89 Hz, 164.44 Hz and 170 Hz, to acquire the corresponding frequency responses. As will be shown later, these frequency points essentially cover the first resonance of the structure when it is subject to uncertainties. Thus, we have $n=1$ and $p=10$ in Equations (3) and (10). In this research, we employ finite element code developed by ourselves using MATLAB to carry out the investigation. This will facilitate a streamlined process for response variation prediction.

Our goal is to accomplish the efficient frequency response variation characterization under uncertainties. For illustration, we divide the structure into 6 segments (Figure 3). We let the mass denisty and the Young's modulus of each segment be subject to variations. Therefore, we have 12 uncertainty parameters and the dimension of $\boldsymbol{\theta}$ shown in Equation (1) is 12. We assume the uncertainty parameters are statistically independent and subject to a multivariate normal distribution with zero means and 10% standard deviations with respect to the nominal values. Although the uncertainties here are parametrized using mass density and Young's modulus, they are reflected in the variations of mass and stiffness matrices of the respective segments. Therefore implicitily a variety of root causes of uncertainties are covered. Using this distribution, 1,000 samples of uncertainty parameters are produced by Latin Hypercube sampling method [33]. These sampled parameters are then employed in full-scale finite element analysis (Section 2.1) and reduced-order analysis (Section 2.2), respectively, to facilitate Monte Carlo simulation to generate the high- and low-fidelity frequency responses. These datasets will be utilized for meta-model training and validation.

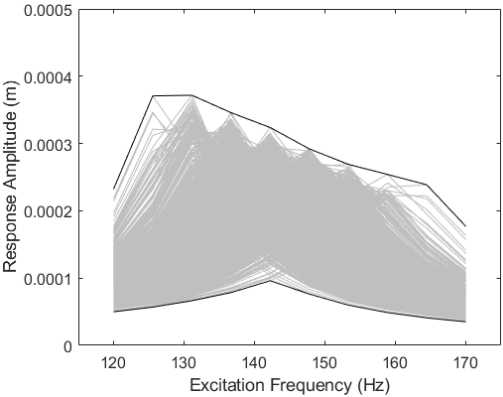

Figure 4. High-fidelity frequency response data (1,000 samples).

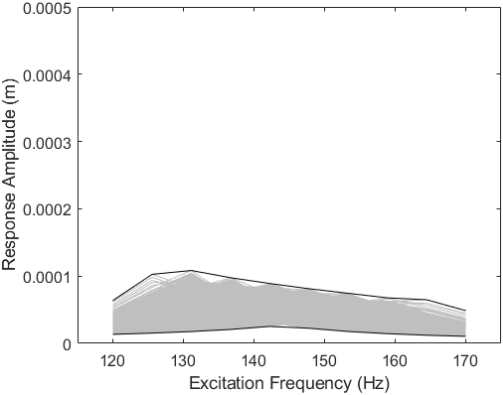

Figure 5. Low-fidelity frequency response data (1,000 samples)

Figures 4 and 5 show the aggregations of 1,000 high- and low-fidelty frequency response data. The frequency range selected, i.e., from 120 Hz to 170 Hz, can cover the resonances of all samples. In both figures, envelops of the upper and lower bounds are included, and the same range of vertical axis is used for clear comparison. The low-fidelity responses have considerable errors when compared to high-fidelity ones in term of response magnitude. We also examine the statistical distribution of high-fidelity response data in Figure 6. Recall that the responses at 10 excitation frequency points are of interest. One may notice that the response distributions under the 4$^{th}$ to 7$^{th}$ excitation frequencies, which are close to the first natural frequency, are quite different from the normal distribution even though the uncertainty parameters are subject to normal distributions. The underlying reason is that, near the natural frequency, the responses are close to being singular and thus are especially sensitive to uncertainties. As a result, the relation between uncertainty parameters (i.e., inputs) and the frequency responses (i.e., outputs) is quite nonlinear. Meanwhile, while some other response distributions under the 3$^{rd}$ and 8$^{th}$ excitation frequencies resemble the normal distribution, there are a plenty of outliers that exist. Apparently, the considerable error in low-fidelity dataset, the nonlinearity observed, and the outliers altogether pose a challenge to developing meta-model.

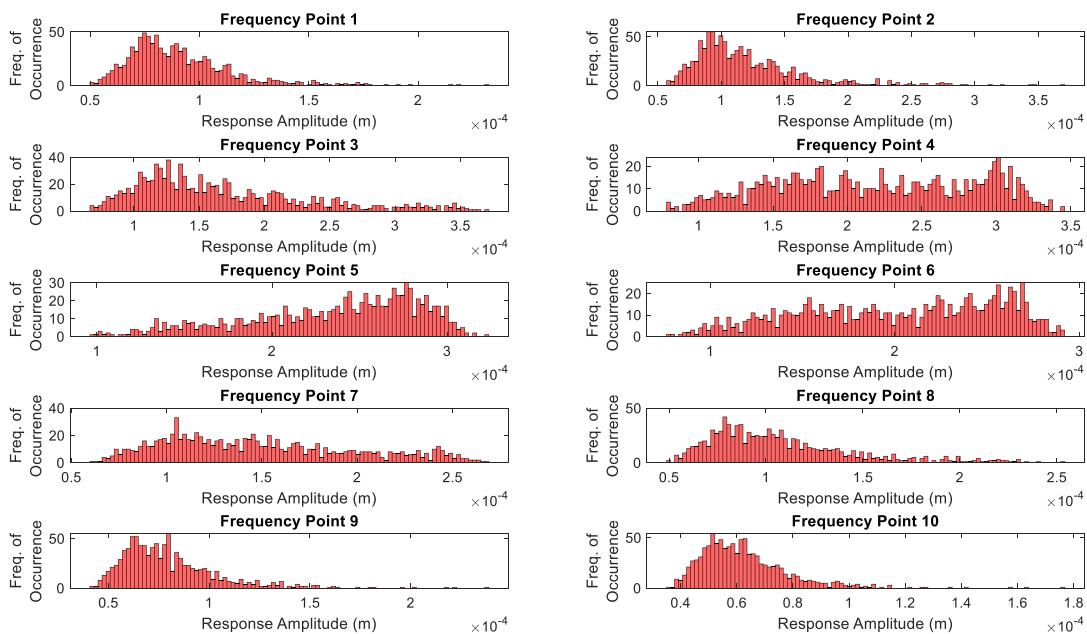

Figure 6. Statistical distributions of high-fidelity frequency response data (1,000 samples).

### *4.2. Meta-model training and implementation details*

The prediciton of frequency response variation requires the development of a learninig approach to be able to emulate responses at different locations and employ concurrently multi-fidelity datasets generalted from different first principle models. We aim at overcoming the limitations of existing approaches including Gaussian processes [22, 34, 35]. As indicated in Section 3.2, we resort to the MFDF-CNN architecture which has the prospect of addressing the implicit, nonlinear relation between the high- and low-fidelity datasets. We further want to examine its performance in the case that the low-fidelity data have considerable error as shown in Section 4.1.

We now re-visit Figure 2 and explain the layout details of MFDF-CNN with the case demonstration. Input layer #1 has 12 nodes, carrying the information of 12 uncertainty parameters as input variables. Hidden layers #2 to #5, each with 512 nodes, are constructed to emulate the relation between the input and the low-fidelity output that is characterized by output layer #5 with 10 nodes. Here, the number of nodes in the output layer is equal to the number of response variables, i.e., response amplitudes evaluated under 10 excitation frequencies of interest. Input layer #1 once again will be concatenated with output layer #5 that are used as new input for high-fidelity output prediction. There are two parallel hidden layer ensembles, #6 to #7, and #8 to #10, that are built respectively to characterize implicitly the linear and nonlinear correlations between the abovementioned new input and the high-fidelity output. As mentioned in Section 3.2, the effect of linearity correlation is considered under weighting coefficient $\alpha$, which is a hyper-parameter to be tuned. Thus, the nonlinear or linear behavior of hidden layers can be simply realized by assigning proper activation function. The layer information and the relevant parameters are listed in Table 2. The total number of weights and biases in the model that need to be optimized is 751,390. By using the loss function defined in Equation (19), MFDF-CNN training can be executed.

Table 2. Layer description and operating parameter set-up in MFDF-CNN

| ID | Layer Description | Size |
|---|---|---|
| #1 | Input layer (dense) | (12,1) |
| #2 | Hidden layer (dense , 'relu' activation) | (512, 1) |
| #3 | Hidden layer (dense , 'relu' activation) | (512, 1) |
| #4 | Hidden layer (dense , 'relu' activation) | (512, 1) |
| #5 | Output layer (dense, 'linear' activation)/low-fidelity | (10, 1) |
| #6 | Hidden layer (dense , no activiation) | (256, 1) |
| #7 | Hidden layer (dense , no activiation) | (256, 1) |
| #8 | Hidden layer (dense , 'relu' activation) | (256, 1) |
| #9 | Hidden layer (dense , relu activiation) | (256, 1) |
| #10 | Hidden layer (dense , relu activiation) | (256, 1) |
| #11 | Output layer (dense, 'linear' activiation) /high-fidelity (linear) | (10, 1) |
| #12 | Output layer (dense, 'linear' activiation) /high-fidelity (nonlinear) | (10, 1) |
| #13 | Output layer (dense, 'linear' activiation) /high-fidelity | (10, 1) |

One major difference between MLMRGP and MFDF-CNN in data usage in the training process exists. As outlined in Section 3.1, MLMRGP allows the independent training of dataset within the same fidelity level, and thus finishes the training of the multi-fidelity datasets sequentially. MFDF-CNN, on the other hand, requires the simultaneous usage of multi-fidelity data points under the same inputs (i.e., sampled uncertainty parameters) for all data points involved. In practice, however, high-fidelity data are usually scarce, and one usually cannot afford to have the same amount of high- and low-fidelity data. Here we introduce an important step to address this issue of data size inconsistency. We expand the size of high-fidelity dataset into the same as the low-fidelity dataset. The missing high-fidelity response data are filled with the corresponding low-fidelity response data under the same uncertainty parameters. In other words, the additional data in the high-fidelity dataset are in fact low-fidelity data. For notation purpose, we refer to them as 'pseudo' high-fidelity data, as shown in Figure 7. We then generate sample weights and assign them for all training samples. To evaluate the loss of one sample, Equation (18) can be re-written as

$$\eta_i = \beta_i^{(1)}\gamma \left\| \mathbf{y}_i^{(1)} - \tau(\mathbf{x}_i) \right\| + \beta_i^{(2)}(1-\gamma) \left\| \mathbf{y}_i^{(2)} - v(\mathbf{x}_i, \tau(\mathbf{x}_i)) \right\| \tag{20}$$

where $\beta_i^{(1)}$ and $\beta_i^{(2)}$ denote, respectively, the $i$-th sample weight values for losses from the low- and high-fidelity outputs. It is noted that the sample weights also will impact the final loss contribution, as in a abroad sense $\beta_i^{(1)}\gamma$ and $\beta_i^{(2)}(1-\gamma)$ literally represent the final loss weights. Specifically, we set sample weights $\beta_i^{(2)}$ as close to zero for the samples coming from the pseudo high-fidelity training data. Such sample weight assignment removes the effect of those pseudo high-fidelity data on training.

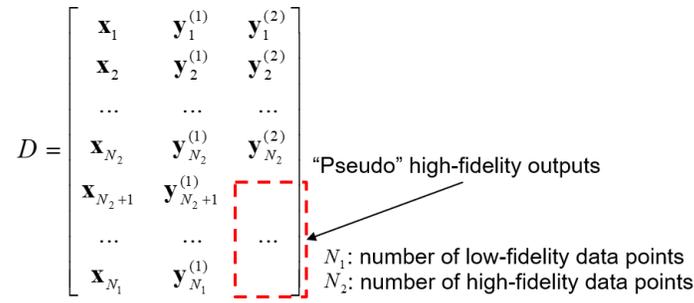

Figure 7. Illustration of treatment of data size inconsistency in MFDF-CNN.

Table 3. Algorithmic setups of MFDF-CNN and MLMRGP

| | MFDF-CNN | MLMRGP |
|---|---|---|
| Data preparation | 40% low-fidelity data (400) and 4% high-fidelity data (40) employed as training datasets. 60% high-fidelity data (600) used as testing dataset – *same data split for both of models* | |
| Operating variables | 1. Epoch size is set as 40. Batch size is set as 5<br>2. $\alpha_i$ and $\gamma_i$ in Equations (17) and (18) are defined as 0.6 and 0.8, respectively<br>3. $\beta_i^{(1)}$ and $\beta_i^{(2)}$ in Equation (20) are selected as 0.5 and 2 respectively for samples with high-fidelity output, and 0.5 and $10^{-5}$ respectively for samples without high-fidelity output | 1. Linear mean kernel<br>2. Anisotropic exponential covariance kernel with 6 reciprocals of lengthscales at each level's emulator plus 1 regression coefficient at high-level's emulator, yielding a total of 13 hyper-parameters |
| Training algorithm | Adam optimizer [36] | Particle swarm optimizer [37] |
| Data preprocessing | Not necessary | Data scaling |

We then proceed to employing the data generated in Section 4.1 to train the MFDF-CNN meta-model. For comparison purpose, the same data are used to train a MLMRGP meta-model which will be subsequently used to elucidate the performance improvement. The same random split of training and testing datasets is applied to both methods. The setup of the two algorithms are listed in Table 3. As indicated in Section 4.1, we generate 1,000 samples of model parameters with uncertainty and subsequently acquire 1,000 high-fidelity data (using finite element analysis) and 1,000 low-fidelity data (using reduced-order analysis). Here for both algorithms, we use 400 low-fidelity data and 40 high-fidelity data for the purpose of training.

*4.3 Frequency response emulation result discussion*

The performance of a meta-model trained can be observed by comparing its emulation result with respect to the actual result obtained through simulation such as finite element analysis. In this research, we're interested in the efficient characterization of frequency response variation induced by model parameter uncertainties. In the data preparation stage, we obtain 1,000 high-fidelity data and 1,000 low-fidelity data. As pointed out in Section 4.2, we use 400 low-fidelity data and 40 high-fidelity data to train the metal models. Recall that in total we have 1,000 sampled model parameters with uncertainty. Each low-fidelity data point corresponds to a specific model parameter sample. We now use those 600 high-fidelity data corresponding to the rest of the model parameter samples for validation/testing.

We compare the emulation results predicted by MLMRGP and MFDF-CNN with respect to the high-fidelity data. In Figure 8 to Figure 17, we plot the frequency response amplitudes versus the testing uncertainty parameter samples, where each figure shows the results comparison under a specific excitation frequency. The frequency response amplitudes shown in Figures 10 to 14 are generally larger, as the corresponding excitation frequencies are closer to the natural frequency. We can readily observe that discrepancies between the testing data and the predicted responses obtained by MFDF-CNN are much smaller than those by MLMRGP, especially for those with large response amplitudes. From the physics perspective, large response occurs in the vicinity of resonance which exhibits the complex relation with respect to the model parameter variation. This is the first indication that MFDF-CNN outperforms MLMRGP in terms of accuracy.

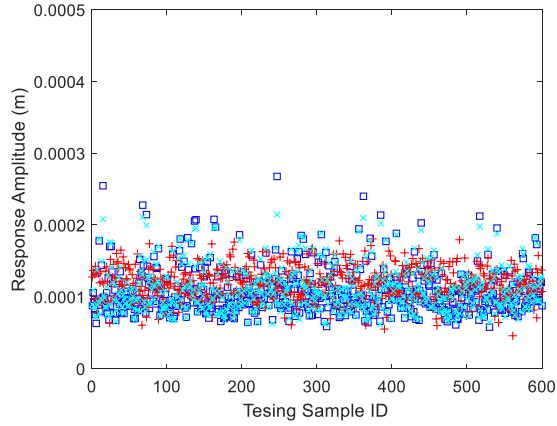

Figure 8. Scatter distribution of response amplitudes under excitation frequency point 1 (120 Hz) over testing space. □: testing/actual outputs; ×: prediction by MFDF-CNN; +: prediction by MLMRGP.

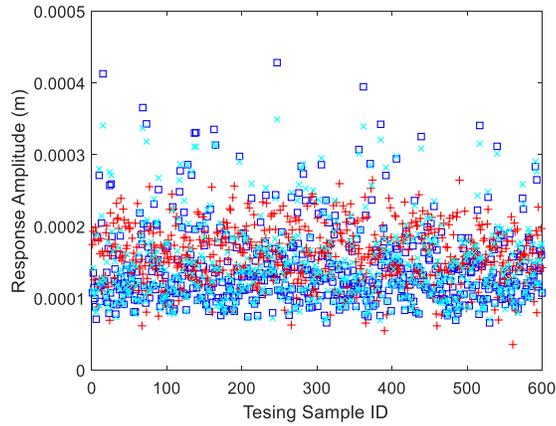

Figure 9. Scatter distribution of response amplitudes under excitation frequency point 2 (125.56 Hz) over testing space. □: testing/actual outputs; ×: prediction by MFDF-CNN; +: prediction by MLMRGP.

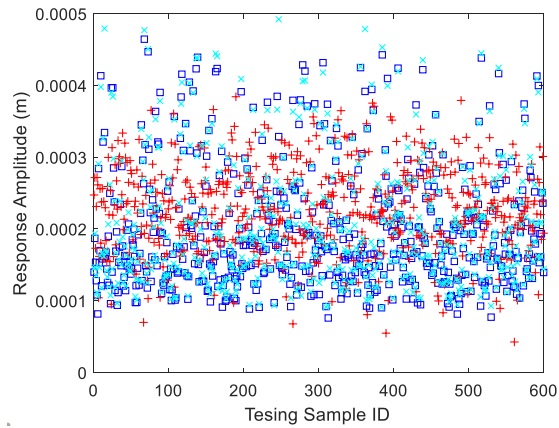

Figure 10. Scatter distribution of response amplitudes under excitation frequency point 3 (131.11 Hz) over testing space. □: testing/actual outputs;×: prediction by MFDF-CNN;+: prediction by MLMRGP.

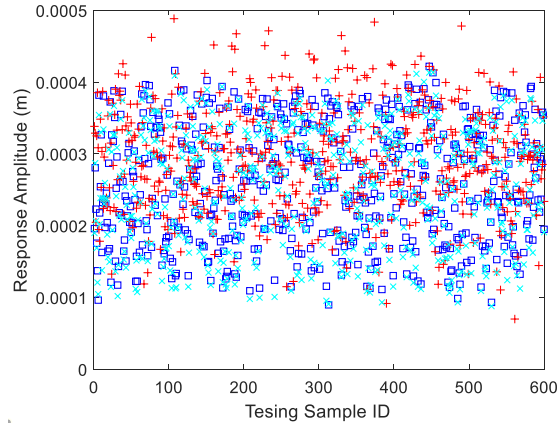

Figure 11. Scatter distribution of response amplitudes under excitation frequency point 4 (136.67 Hz) over testing space. □: testing/actual outputs;×: prediction by MFDF-CNN;+: prediction by MLMRGP.

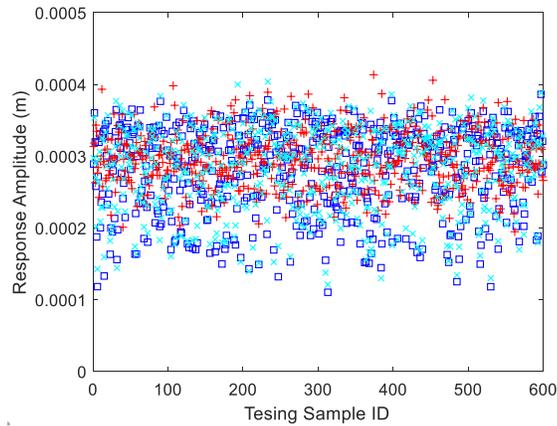

Figure 12. Scatter distribution of response amplitudes under excitation frequency point 5 (142.22 Hz) over testing space. □: testing/actual outputs;×: prediction by MFDF-CNN;+: prediction by MLMRGP.

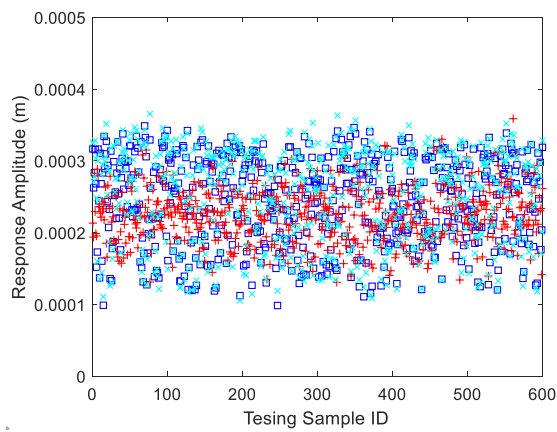

Figure 13. Scatter distribution of response amplitudes under excitation frequency point 6 (147.78 Hz) over testing space. □: testing/actual outputs;×: prediction by MFDF-CNN;+: prediction by MLMRGP.

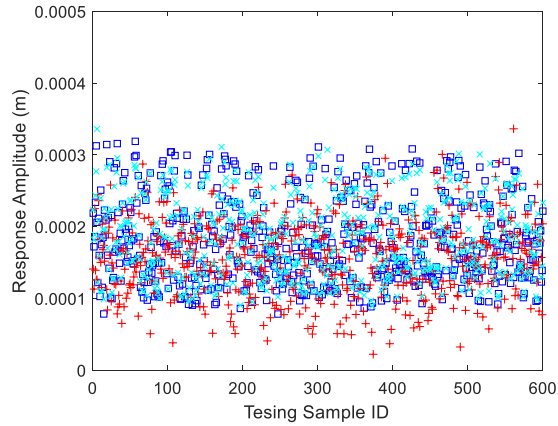

Figure 14. Scatter distribution of response amplitudes under excitation frequency point 7 (153.33 Hz) over testing space. □: testing/actual outputs; ×: prediction by MFDF-CNN; +: prediction by MLMRGP.

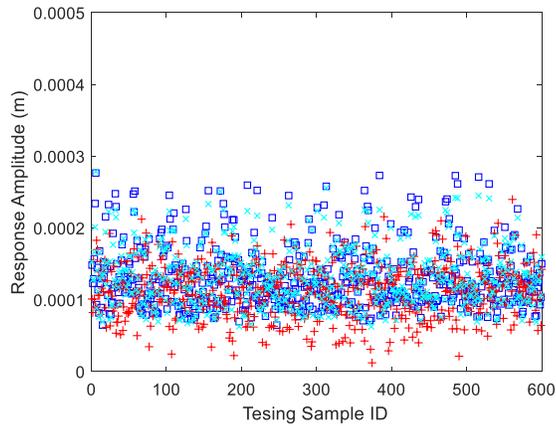

Figure 15. Scatter distribution of response amplitudes under excitation frequency point 8 (158.89 Hz) over testing space. □: testing/actual outputs; ×: prediction by MFDF-CNN; +: prediction by MLMRGP.

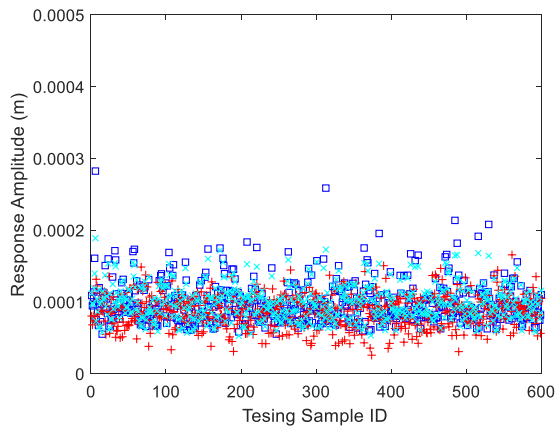

Figure 16. Scatter distribution of response amplitudes under excitation frequency point 9 (164.44 Hz) over testing space. □: testing/actual outputs; ×: prediction by MFDF-CNN; +: prediction by MLMRGP.

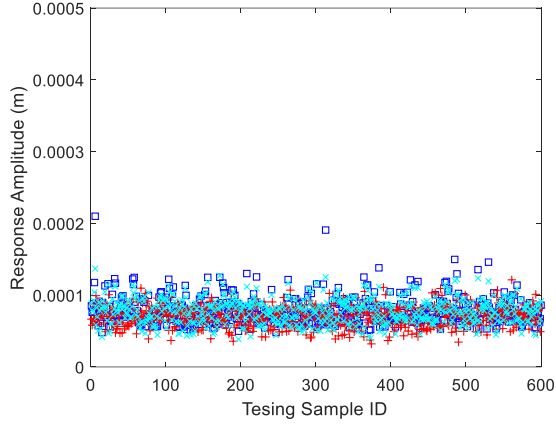

Figure 17. Scatter distribution of response amplitudes under excitation frequency point 10 (170 Hz) over testing space. □: testing/actual outputs; ×: prediction by MFDF-CNN; +: prediction by MLMRGP.

To quantify the comparison, here we also employ the mean squared error (MSE) between the emulation result and the actual result as

$$\eta_r^* = \frac{1}{M} \sum_{k=1}^{M} \eta_{k,r}^2 \tag{21}$$

where

$$\eta_{k,r}^* = \left| \mathbf{y}_{k,r}^{(2)*} - \hat{v}(\mathbf{x}_{k,r}, \hat{\tau}(\mathbf{x}_{k,r})) \right| \tag{22}$$

In above equations, the subscripts $k$ and $r$ indicate the $k$-th testing sample and the $r$-th excitation frequency, respectively. For example, the vector $\mathbf{y}_{k,r}^{(2)*}$ represents the $r$-th frequency response amplitudes of the $k$-th testing sample. $\hat{\tau}$ and $\hat{v}$ represent the well-trained low-and high-level emulators in MFDF-CNN, respectively. $M$ is the number of testing samples and is 600 in this analysis. It is worth noting that in the training process, all frequency responses are introduced to formulate the mean squared error loss function as shown in Equation (18) or (20) which takes into consideration the weights of different outputs. MSE defined in Equation (21) is different, and intends to measure the difference between the testing and the predicted outputs. It is used to evaluate the prediction accuracy of frequency response at each frequency point over the entire testing samples. The comparison of MSE values is shown in Table 4. In general, MFDF-CNN yields smaller MSE values than MLMRGP. The improvement is more significant at frequencies close to the natural frequency. This indicates that MFDF-CNN can deal with high sensitivity of uncertainty parameters. It is still worth mentioning that there is subtle difference between MFDF-CNN and MLMRGP in terms of optimization objective. Training of MLMRGP essentially follows the Bayesian framework, which aims at maximizing the marginal likelihood upon the training data. On the other hand, MFDF-CNN, as one regression neural network, allows one to adopt different loss functions, such as mean squared error, mean squared logarithmic error, and mean absolute error, etc. One may argue that MLMRGP is subject to a somewhat different optimization objective. Nevertheless, from prediction accuracy standpoint, MFDF-CNN indeed leads to reduced error as a whole.

Table 4. Comparison of MSE values ($m^2$) between MFDF-CNN and MLMRGP

| Frequency Point | MFDF-CNN | MLMRGP |
| --- | --- | --- |
| 1 | $1.3928\times10^{-11}$ | $9.2191\times10^{-10}$ |
| 2 | $3.6517\times10^{-11}$ | $2.7928\times10^{-9}$ |
| 3 | $7.9976\times10^{-11}$ | $6.2554\times10^{-9}$ |
| 4 | $9.9773\times10^{-11}$ | $6.5601\times10^{-9}$ |
| 5 | $8.3908\times10^{-11}$ | $2.1034\times10^{-9}$ |
| 6 | $1.1425\times10^{-10}$ | $2.4425\times10^{-9}$ |
| 7 | $8.0084\times10^{-11}$ | $3.1576\times10^{-9}$ |
| 8 | $2.7335\times10^{-11}$ | $1.7564\times10^{-9}$ |
| 9 | $3.0455\times10^{-11}$ | $7.2190\times10^{-10}$ |
| 10 | $1.9315\times10^{-11}$ | $3.0172\times10^{-10}$ |

Discussed above are the frequency response values at individual excitation frequency points. We also randomly select frequency response curves of 6 testing samples for comparison, as shown in Figure 18. It can be seen that the low-fidelity frequency response curves contain large errors. Both MLMRGP and MFDF-CNN, after incorporating high-fidelity dataset for meta-model training, perform better than the low-fidelity data as the predictions results are pushed towards the corresponding high-fidelity responses (i.e., actual responses). Nevertheless, MFDF-CNN outperforms MFMRGP, as the frequency response curves it predicts basically overlap those of the actual values.

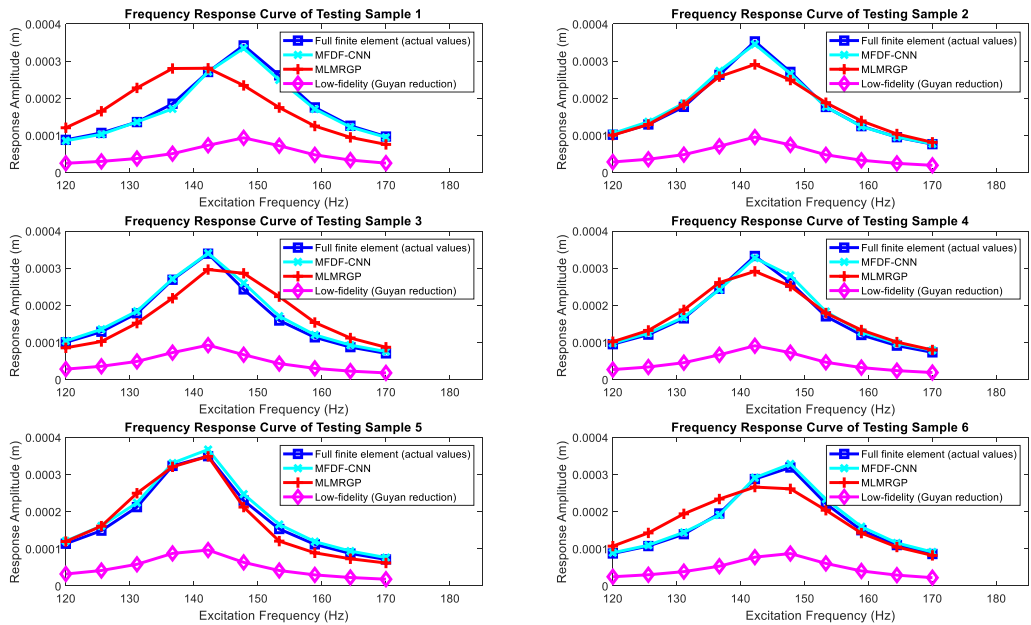

Figure 18. Frequency response curve comparison of 6 selected samples.

While this research primarily focuses on meta-model training accuracy, training efficiency is still one important aspect. Here MFDF-CNN and MLMRGP are implemented at the same computational platform, i.e., Intel CPU E5-2640 @2.40GHz (2 processors). We employ the deep learning library, i.e., Keras written in Python, to develop MFDF-CNN because it is powerful. Meanwhile, we use MATLAB for MLMRGP meta-modeling as it provides efficient matrix operation functions involved in the training of MLMRGP. For each single run, MFDF-CNN and MLMRGP take 57 and 483 seconds, respectively. While these two models are established upon different integrated development environments (IDEs), i.e., Python and MATLAB, MFDF-CNN appears to be more efficient for the case investigated in this research. Even though MFDF-CNN involves a large number of weights and biases to be trained, it can take advantage of the built-in Adam optimizer [36]. MLMRGP on the other hand includes a number of non-sparse matrix operations in the training process, which leads to higher computational cost. Future research may investigate situations where larger amount of training data and larger number of response variables are considered, and conduct a more rigorous comparison of these algorithms under the same integrated development environment (IDE).

### *4.4. Performance robustness and parametric influence*

The accuracy of meta-model, strictly speaking, is subject to certain randomness, due to the random split of training and testing datasets as well as the training process. In the case of neural network, training is facilitated by optimizing the model parameters, i.e., weights and biases, which is influenced by the initial parameter guess, stochastic or gradient-based parameter search, and random training batch generation at each epoch/iteration. To examine the performance robustness, we carry out multiple runs of training and summarize the results statistically. Here we implement 5 runs with random training and testing data splits. In each run, we use 400 low-fidelity and 40 high-fidelity data for training and then use the rest 600 high-fidelity data for testing, which is the same configuration used in Section 4.3. In each run, the same split of training and testing datasets is adopted for both MFDF-CNN and MLMRGP. While we still resort to the MSE defined in Equation (21) as metrics, we take the logarithm for the convenience of illustration, i.e.,

$$L_{t,r}^* = \ln(\eta_r^*) \tag{23}$$

And the mean logarithm MSE is defined as

$$L_r^* = \frac{1}{Q}\sum_{t=1}^{Q} L_{t,r}^* \tag{24}$$

In the above equations, $r$ indicates the $r$-th excitation frequency point, and $t$ indicates the $t$-th training/emulation run under the $t$-th dataset split. $Q$ is total number of emulation runs/dataset splits, which is 5 in this case. The smaller $L_r^*$ is, the better accuracy the meta-model has.

Figures 19 and 20 show the values of defined metric, logarithm MSE $L_{t,r}^*$, under different training/emulation runs using MFDF-CNN and MLMRGP. In both methods, the values of $L_{t,r}^*$ slightly differ under different training/emulation runs. This indicates that both methods perform in a robust manner without obvious accuracy difference. Both methods do not exhibit overfitting. The logarithm values of MSE $L_{t,r}^*$ of MFDF-CNN are consistently smaller than those of MLMRGP. The mean logarithm MSE $L_r^*$ of MFDF-CNN and that of MLMRGP are compared in Figure 21. As can be observed, the accuracy of MFDF-CNN is significantly better than that of MLMRGP.

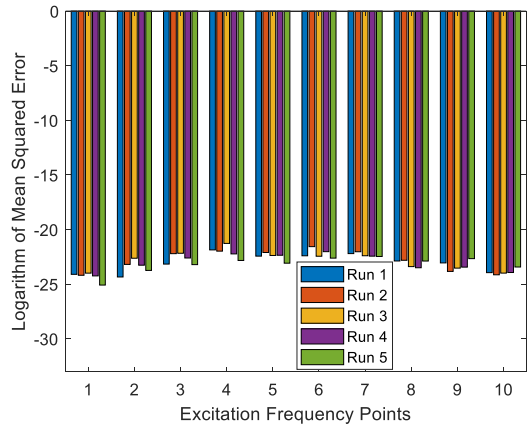

Figure 19. Error expressed as $L_{t,r}^*$ of multiple training/emulation runs through MFDF-CNN.

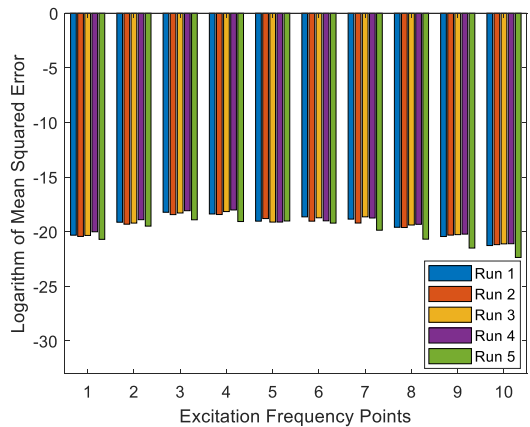

Figure 20. Error expressed as $L_{t,r}^*$ of multiple training/emulation runs through MLMRGP.

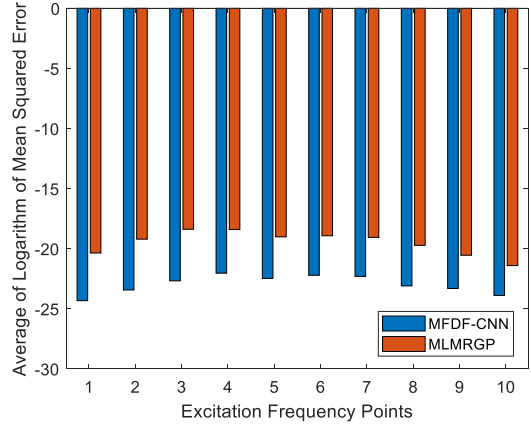

Figure 21. Comparison of average error expressed as $L_r^*$ under MFDF-CNN and MLMRGP.

In multi-fidelity data fusion, the error of low fidelity data is inherited from the order-reduction procedure, and we hope to correct such error through data fusion by introducing a few high-fidelity data. In Figure 18, we can observe such positive influence of the high-fidelity data. In meta-model training, in general, more training data yields better performance in terms of accuracy. Specifically, in multi-fidelity data fusion, intuitively, introducing more high-fidelity data while maintaining the amount of low-fidelity data is expected to further improve the prediction accuracy. Here we examine how the size of high-fidelity training datasets affects the prediction accuracy. In the abovementioned analyses, we use 400 low-fidelity data and 40 high-fidelity data. This amount of high-fidelity data corresponds to 10% of the low-fidelity data. We now increase the size of high-fidelity data to 20% and 30% of the low-fidelity data, and respectively train the meta-models. The logarithm MSE values at the first 5 frequency points under the two meta-modeling approaches are shown in Figure 22. The results of the other 5 frequency points that are essentially symmetric with respect to the first 5 along the first natural frequency exhibit similar trends. As can be observed, increasing high-fidelity dataset leads to the performance improvement. Interestingly, it can be observed that MFDF-CNN always outperforms MLMRGP even as the high-fidelity dataset size increases.

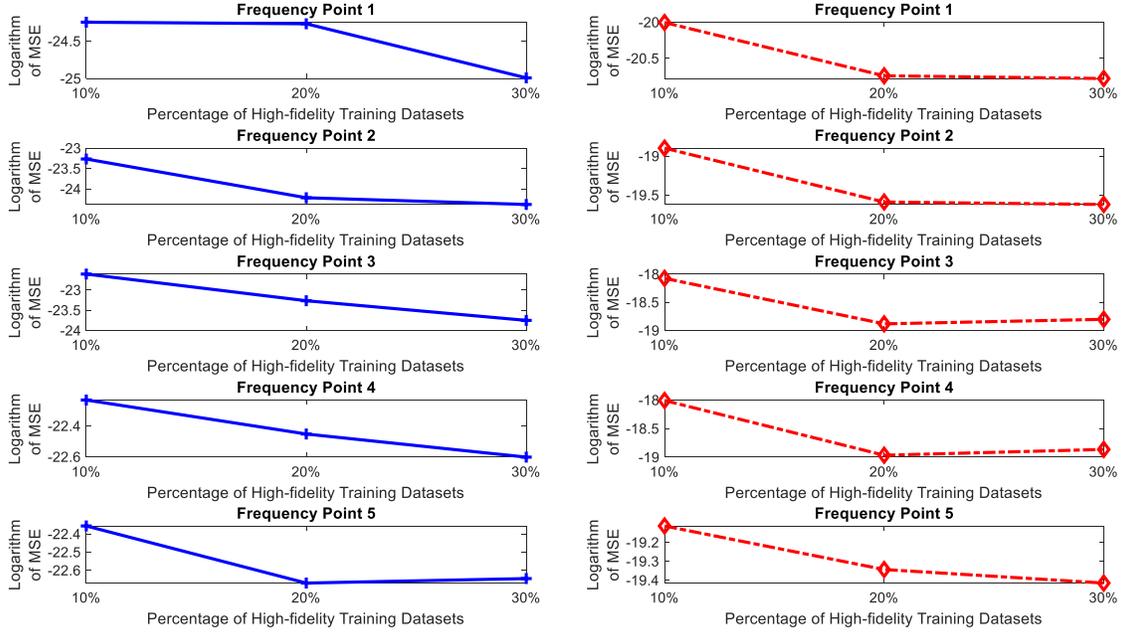

Figure 22. Comparison of prediction errors with respect to high-fidelity training data size. +: result of MFDF-CNN; ◇: result of MLMRGP.

One advantage of MFDF-CNN, as compared to MLMRGP, is that it can take implicitly both linear and nonlinear correlations of multi-fidelity outputs into account. As shown in Equation (17), the effects of linear and nonlinear correlations can be quantitatively represented by $\alpha$ and $1-\alpha$. Observe the frequency response curves shown in Figure 18. The high-fidelity result and the low-fidelity result exhibit more significant difference near the natural frequency. Intuitively, there is more significant nonlinear correlation at frequency points near the natural frequency. Here we specifically investigate the effect of linearity weight $\alpha$ (Equation (17)). As mentioned, $\alpha$ is a hyper-parameter of MFDF-CNN meta-model, which cannot be directly identified through training. Here, we apply the uniform-grid discretization of $\alpha$ within range [0, 1] with 0.1 increment, and evaluate the errors (i.e., $L_{t,r}^*$ in Equation (23a)) of all $\alpha$ values given the same split of training and testing datasets. The result shown in Figure 23 indicates that the errors vary with respect to $\alpha$. In Section 4.3, we adopt $\alpha = 0.6$ for the MFDF-CNN meta-modeling. In Figure 23, it is found that $\alpha = 0.6$ indeed yields a relatively small error. In other words, there exists considerable nonlinear correlations between the low-fidelity and high-fidelity datasets. As compared with other techniques such as MLMRGP that are built upon the linear relation assumption, this MFDF-CNN allows the characterization of more generic relation of multi-fidelity datasets. As a result, it exhibits much enhanced predictive capability.

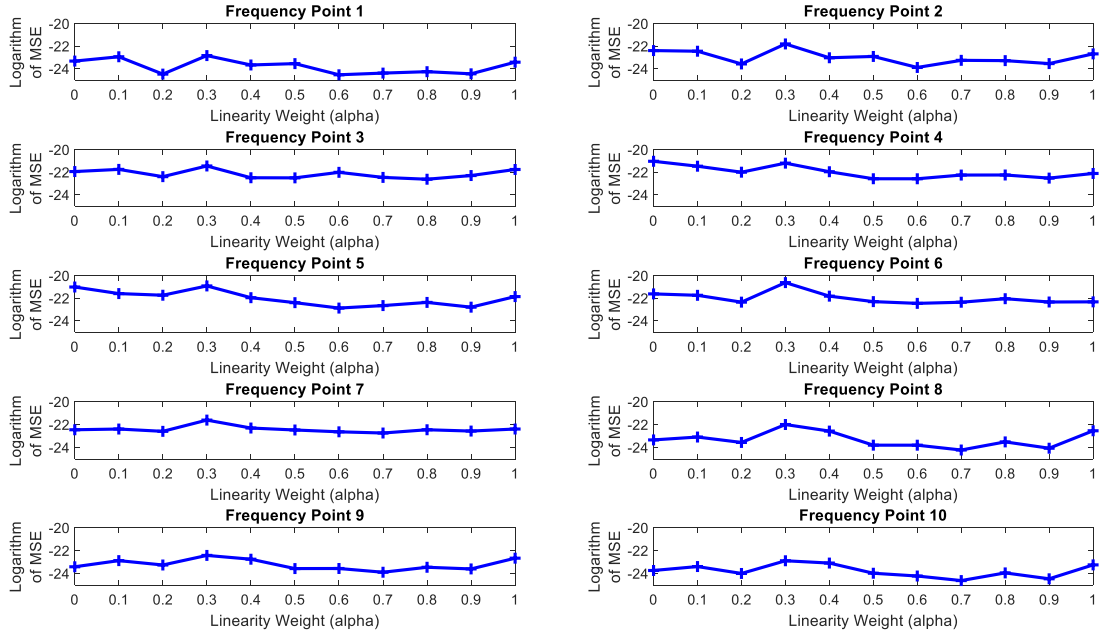

Figure 23. Prediction error with respect to linearity weight.

## 4. Conclusion

In this research, a multi-fidelity data fusion composite neural network (MFDF-CNN) is developed to efficiently and accurately characterize the frequency response variation of a structure under uncertainties. This approach, by taking advantage of the architecture of composite neural network, allows one to integrate together a small amount of high-fidelity data acquired from full-scale finite element analysis and a large amount of low-fidelity data acquired from reduced-order analysis. This can significantly reduce the computational cost of data acquisition for the subsequent meta-model training and validation. Moreover, this approach enables the characterization of the implicit relation between low- and high-fidelity datasets, which can yield the improved accuracy as compared with the state-of-the-art techniques. The case studies demonstrate that this multi-fidelity data fusion approach can effectively improve the variation prediction capability with performance robustness.

## Acknowledgment

This research is supported by NSF under grant CMMI – 1741174.

## References

[1] Mao, Z., Todd, M. Statistical modeling of frequency response function estimation for uncertainty quantification. Mechanical Systems and Signal Processing 2013; 38(2):333-345.

[2] Brehm, M., Deraemaeker, A. Uncertainty quantification of dynamic responses in the frequency domain in the context of virtual testing. Journal of Sound and Vibration 2015; 342:303-329.

[3] Stefanou, G. The stochastic finite element method: past, present and future. Computer Methods in Applied Mechanics and Engineering 2009; 198(9-12):1031-1051.

[4] Sofi, A., Romeo, E. A unified response surface framework for the interval and stochastic finite element analysis of structures with uncertain parameters. Probabilistic Engineering Mechanics 2018; 54:25-36.

[5] Zhou, K., Hegde, A., Cao, P., Tang, J. Design optimization towards alleviating forced response variation in cyclically periodic structure using Gaussian process. Journal of Vibration and Acoustics, Transactions of the ASME 2017;139(1): 011017.

[6] Fricker, T.E., Oakley, J.E., Sims, N.D., Worden, K. Probabilistic uncertainty analysis of an FRF of structure using a Gaussian process emulator. Mechanical Systems and Signal Processing 2011; 25(8):2962-2975.

[7] Hariri-Ardebili, M.A., Sudret, B. Polynomial chaos expansion for uncertainty quantification of dam engineering problems. Engineering Structures 2020; 203:109631.

[8] Tripathy, R.K., Bilionis, I. Deep UQ: Learning deep neural network surrogate models for high dimensional uncertainty quantification. Journal of Computational Physics 2018; 375:565-588.

[9] Arendt, P. D., Apley, D. W., Chen, W., Lamb, D., Gorsich, D. Improving Identifiability in Model Calibration Using Multiple Responses. Journal of Mechanical Design, Transactions of the ASME 2012; 134(10):100909.

[10] Arendt, P. D., Apley, D. W., Chen, W. Quantification of Model Uncertainty: Calibration, Model Discrepancy, and Identifiability. Journal of Mechanical Design, Transactions of the ASME 2012; 134(10): 100908.

[11] Wan, H.P., Ni, Y.Q. Bayesian multi-task learning methodology for reconstruction of structural health monitoring data. Structural Health Monitoring, 2019; 18:1282-1309.

[12] Hassoun, M.H. *Fundamentals of artificial neural networks*. MIT Press, 1995.

[13] AI-Momani, E.S., Mayyas, A.T., Rawabdeh, I., Alqudah, R. Modeling blanking process using multiple regression analysis and artificial neural networks. Journal of Materials Engineering and Performance 2012; 21:1611-1619.

[14] Du, D., Li, K., Fei, M. A fast multi-output RBF neural network construction method. Neurocomputing 2010; 73(10-12):2196-2202.

[15] Lee, S.C. Prediction of concrete strength using artificial neural networks. Engineering Structures 2003; 25(7):849-857.

[16] Min, J., Park, S., Yun, C.B., Lee, C.G., Lee, C. Impedance-based structural health monitoring incorporating neural network technique for identification of damage type and severity. Engineering Structures 2012; 39:210-220.

[17] Park, Y.S., Kim, S., Kim, N., Lee, J.J. Finite element model updating considering boundary conditions using neural networks. Engineering Structures 2017; 150:511-519.

[18] Kennedy, M.C., O'Hagan, A. Predicting the output from a complex computer code when fast approximation are available. Biometrika 2000; 87(1):1-13.

[19] Raissi, M., Perdikaris, P., Karniadakis, G.E. Inferring solutions of differential equations using noisy multi-fidelity data. Journal of Computational Physics 2017; 335:736-746.

[20] Liu, H.T., Ong, Y.S., Cai, J.F., Wang, Y. Cope with diverse data structures in multi-fidelity modeling: A Gaussian process method. Engineering Applications of Artificial Intelligence 2018; 67:211-225.

[21] De Lima, A.M.G., Da Silva, A.R., Rade, D,A., Bouhaddi, N. Component mode synthesis combining robust enriched Ritz approach for viscoelastically damped structures. Engineering Structures 2010; 32(5): 1479-1488.

[22] Zhou, K.., Tang, J. Uncertainty quantification in structural dynamic analysis using two-level Gaussian processes and Bayesian inference. Journal of Sound and Vibration 2018; 412:95-115.

[23] Zhou, K., Tang, J. Uncertainty Quantification of Mode Shape Variation Utilizing Multi-Level Multi-Response Gaussian Process. ASME Journal of Vibration and Acoustics. In revision (arXiv:2002.09287). .

[24] Perdikaris, P., Raissi, M., Damianou, A., Lawrence, N., Karniadakis, G.E. Non-linear information fusion algorithms for data-efficient multi-fidelity modelling. Mathematical, Physical and Engineering Sciences 2017; 473(2198):20160751.

[25] Meng, X.H., Karniadakis, G.E. A composite neural network that learns from multi-fidelity data: Application to function approximation and inverse PDE problems. Journal of Computational Physics 2020; 401:109020.

[26] Craig, R.R., Kurdila, A.J., *Fundamentals of structural dynamic*, Wiley, 2006.

[27] Basheer, I.A., Hajmeer, M. Artificial neural networks: fundamentals, computing, design, and application. Journal of Microbiological Methods 2000; 43:3-31.

[28] Cao, P., Zhang, S.L., Tang, J. Preprocessing-free gear fault diagnosis using small datasets with deep convolutional neural network-based transfer learning. IEEE Access 2018; 6:26241-26253.

[29] Zhang, W., Li, C.H., Peng, G.L., Chen, Y.H., and Zhang, Z.J. A deep convolutional neural network with new training methods for bearing fault diagnosis under noisy environment and different working load. Mechanical Systems and Signal Processing 2018; 100:439-453.

[30] Hoang, D.T., Kang, H.J. A survey on deep learning based bearing fault diagnosis. Neurocomputing 2019; 335:327-335.

[31] Haykin, S. *Neural Networks: A Comprehensive Foundation (2$^{nd}$ edition)*. Prentice Hall, 1998.

[32] Dangeti, P. *Statistics for Machine Learning*. Packt Publishing Ltd, 2017.

[33] Han, Z.H., Gortz, S. Hierarchical kriging model for variable-fidelity surrogate modeling. AIAA Journal 2012; 50:1885-1896.

[34] Gratiet, L.L., Garnier, J. Recursive co-kriging model for design of computer experiments with multiple levels of fidelity. International Journal for Uncertainty Quantification 2014; 4:365-386.

[35] Kroese, D.P., Taimre, T., Botev, Z.I. *Handbook of Monte Carlo Methods*. Wiley, 2011.

[36] Kingma, D.P., Ba, J. Adam: A method for stochastic optimization. arXiv:1412.6980 [cs.LG] 2017.

[37] Clerc, M. *Particle Swarm Optimization*. Wiley Online Library, 2006.